\documentclass{article}

\PassOptionsToPackage{numbers, square}{natbib}

\usepackage[preprint]{neurips_2025}

\usepackage[utf8]{inputenc} 
\usepackage[T1]{fontenc}    
\usepackage{hyperref}       
\usepackage{url}            
\usepackage{booktabs}       
\usepackage{amsfonts}       
\usepackage{nicefrac}       
\usepackage{microtype}      
\usepackage{xcolor}         
\usepackage{graphicx}       
\usepackage{algorithm}      
\usepackage{algorithmic}    
\usepackage{float}          
\usepackage{placeins}       
\usepackage{multirow}       

\title{A Foundation Chemical Language Model for Comprehensive Fragment-Based Drug Discovery}

\author{%
    Alexander Ho$^{1,2}$, Sukyeong Lee$^{1,2}$, Francis T.F. Tsai$^{2,3,4}$\\
    $^{1}$Advanced Technology Core for Macromolecular X-Ray Crystallography\\
    $^{2}$Verna and Marrs McLean Dept. of Biochemistry and Molecular Pharmacology\\
    $^{3}$Dept. of Molecular \& Cellular Biology\\
    $^{4}$Dept. of Molecular Virology \& Microbiology\\
    Baylor College of Medicine\\
    Houston, TX 77030 \\
    	\texttt{alexander.ho@ucsf.edu\thanks{Corresponding author, present address}, slee@bcm.edu, ftsai@bcm.edu\thanks{Corresponding author}} \\
}
\begin{document}

\maketitle
\begin{abstract}
We introduce FragAtlas-62M, a specialized foundation model trained on the largest fragment dataset to date. Built on the complete ZINC-22 fragment subset comprising over 62 million molecules, it achieves unprecedented coverage of fragment chemical space.
Our GPT-2 based model (42.7M parameters) generates 99.90\% chemically valid fragments. Validation across 12 descriptors and three fingerprint methods shows generated fragments closely match the training distribution (all effect sizes < 0.4). The model retains 53.6\% of known ZINC fragments while producing 22\% novel structures with practical relevance. 
We release FragAtlas-62M with training code, preprocessed data, documentation, and model weights to accelerate adoption.
\end{abstract}

\begin{figure}[t]
\centering
\FloatBarrier
\includegraphics[width=0.95\textwidth]{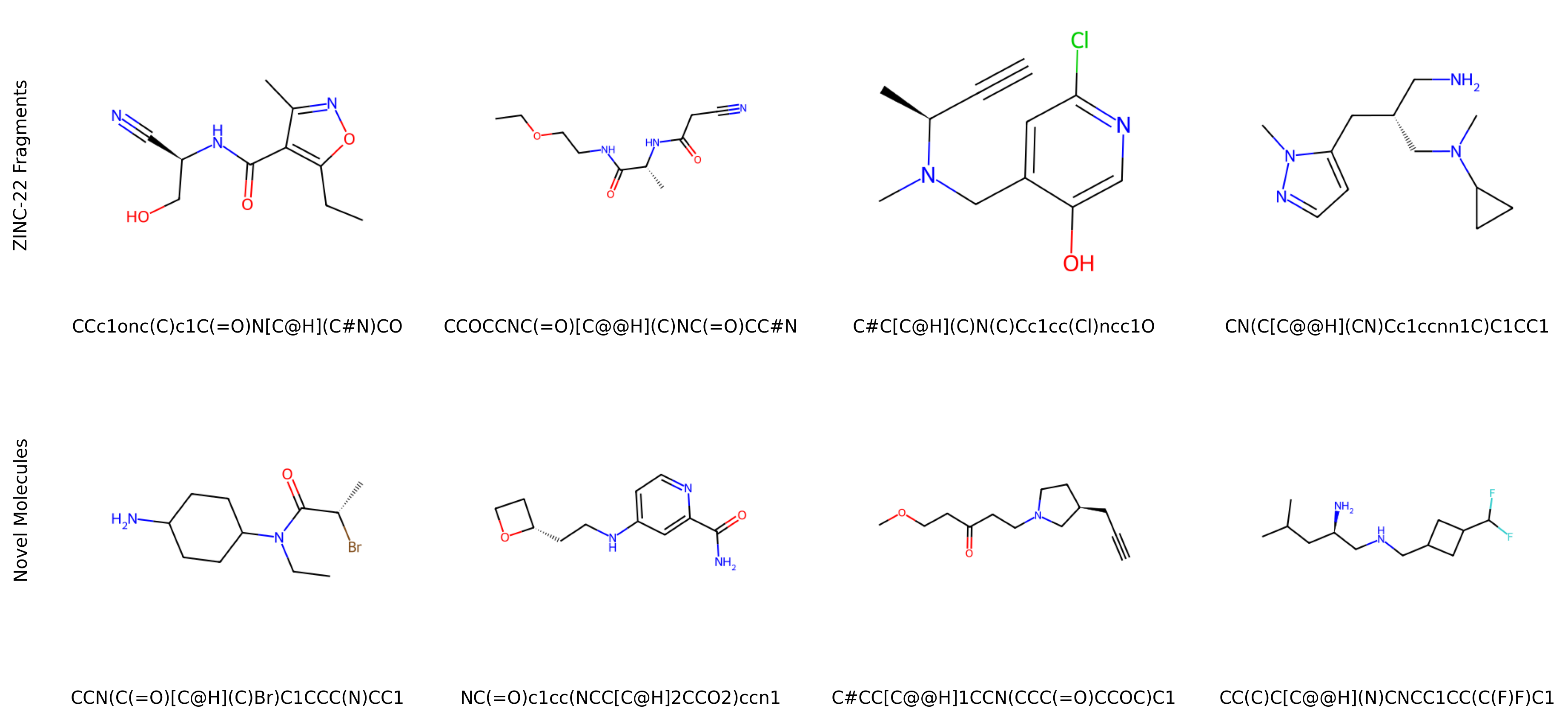}
\caption{Representative molecules from ZINC-22 (top) and novel samples from FragAtlas-62M (bottom), shown with SMILES labels. Note: Different molecular visualization tools may use distinct coordinate generation algorithms to convert SMILES strings into 2D structures, resulting in chemically equivalent but visually different representations of the same molecule.}
\label{fig:molecule_comparison}
\end{figure}

\section{Introduction}
Fragment-based drug discovery (FBDD) is a growing approach in medicinal chemistry for identifying low-molecular-weight compounds that serve as starting points for lead optimization \cite{xu2025fragment, keseru2016design}. Despite its importance, existing AI-driven molecular generators primarily target full-sized drug molecules \cite{elton2019deep, ross2022large}, while fragment-specific methods remain limited in scale and scope \cite{bian2018computational}. This leaves a significant gap: no large-scale generative foundation model exists specifically for molecular fragments.

We introduce FragAtlas-62M, a GPT-2 based chemical language model trained on the complete ZINC-22 fragment subset (62,015,589 molecules) \cite{tingle2023zinc}. FragAtlas-62M achieved 99.90\% chemical validity while maintaining 53.55\% coverage of known ZINC fragments and generating 22.04\% novel structures; it preserves the training data's property distributions across multiple descriptors. We release the model, training code, preprocessed data, and documentation openly to the community.

\section{Related Work}

\label{sec:related_work}

\subsection{Chemical Language Models}

Several chemical language models directly relevant to molecular generation have been proposed. MolGPT \cite{bagal2021molgpt} adapts GPT architectures for molecule generation, ChemBERTa \cite{chithrananda2020chemberta} applies BERT-style pretraining to molecular representations, and FragGPT \cite{yue2024unlocking} explores unordered chemical language for molecular design. These models primarily target full-sized drug molecules rather than fragments.

Recent work has also examined how non-canonical or invalid SMILES can act as data augmentation, improving model robustness \cite{skinnider2024invalid}; FragAtlas-62M preserves non-canonical SMILES during training and enforces validity only at evaluation to capture this benefit.

\subsection{Fragment-Based Drug Discovery}

Fragment-based drug discovery (FBDD) has emerged as a promising approach in medicinal chemistry \cite{xu2025fragment, keseru2016design}, but computational tools specifically designed for fragment generation remain limited. FBDD offers several advantages over traditional high-throughput screening, including more efficient exploration of chemical space and higher hit rates for challenging targets. The ZINC database \cite{tingle2023zinc} is a key resource, providing commercially accessible molecules from billion-scale libraries, including fragment collections available for experimental validation.

While computational methods have been applied to fragment-based approaches \cite{bian2018computational}, they have largely focused on virtual screening or fragment linking rather than de novo generation. Some studies have explored using generative models with fragments as building blocks \cite{powers2023geometric, podda2020deep}, fragment-based sequential translation \cite{chen2021fragment}, or fragment retrieval augmentation \cite{lee2024molecule}. Frag2Seq \cite{fu2024fragment} offers a promising approach that combines fragment tokenization with geometric information for structure-based design, while FragGPT \cite{yue2024unlocking} leverages unordered chemical language for comprehensive molecular design. However, none of these approaches have developed foundation models that comprehensively address the specific needs of fragment-based approaches at the scale of the entire ZINC fragment database.

\begin{figure}[t]
\centering
\FloatBarrier
\begin{tabular}{@{}c@{}ccc@{}}
& \small{(A) QED} & \small{(B) LogP} & \small{(C) Molecular Weight} \\
\raisebox{1.5cm}{\rotatebox{90}{\textbf{Density}}} &
\includegraphics[width=0.30\textwidth]{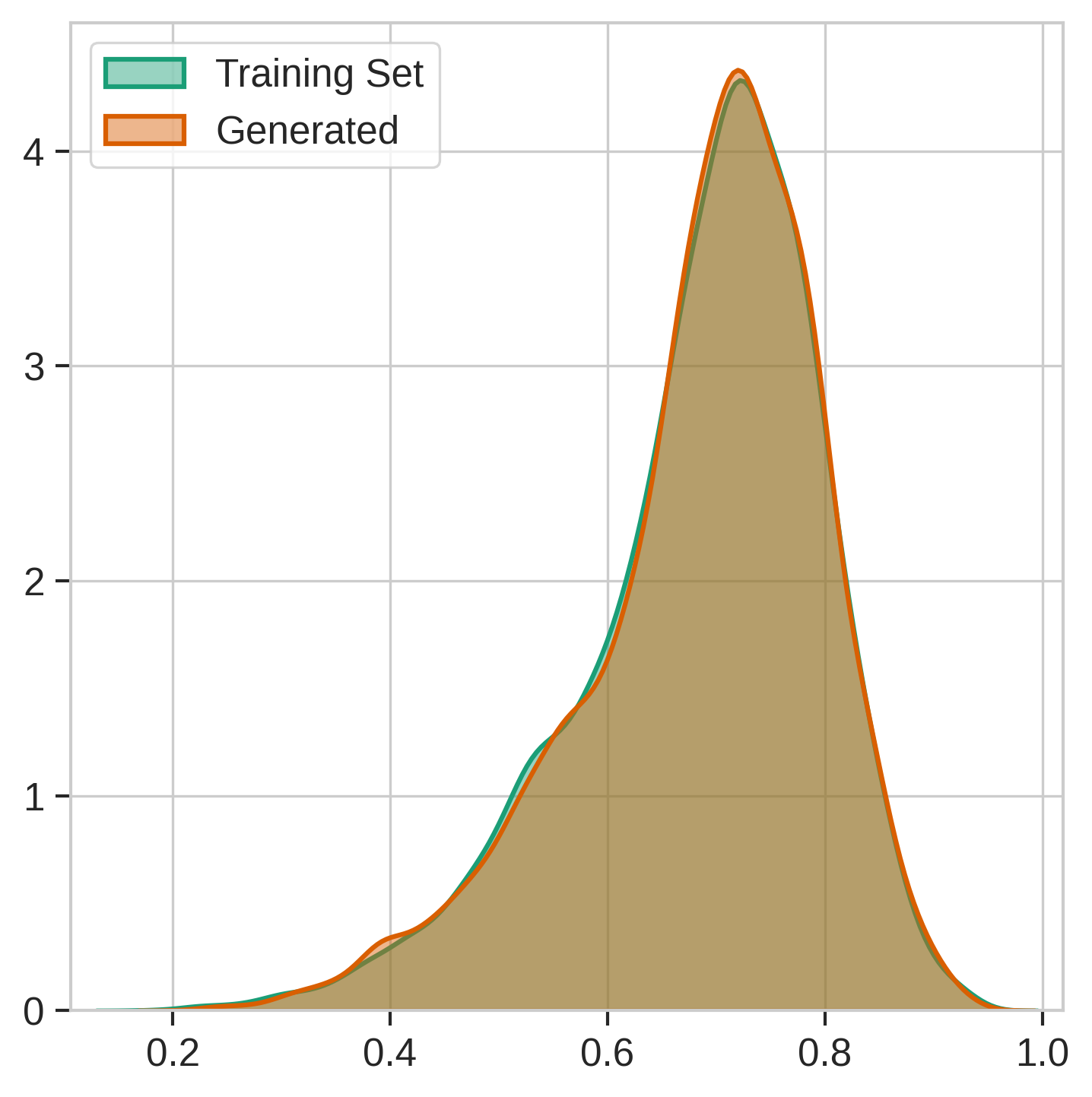} &
\includegraphics[width=0.30\textwidth]{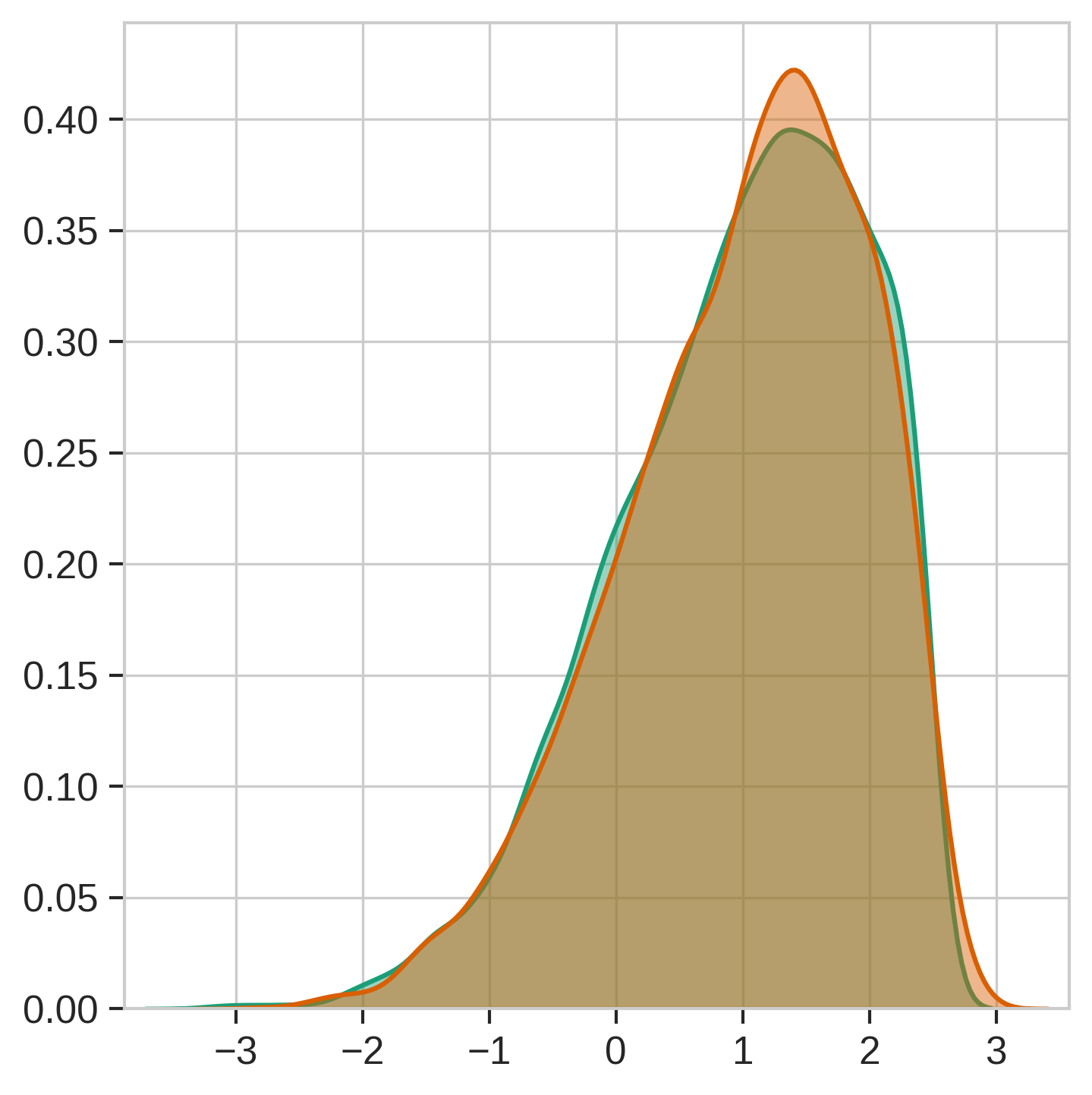} &
\includegraphics[width=0.30\textwidth]{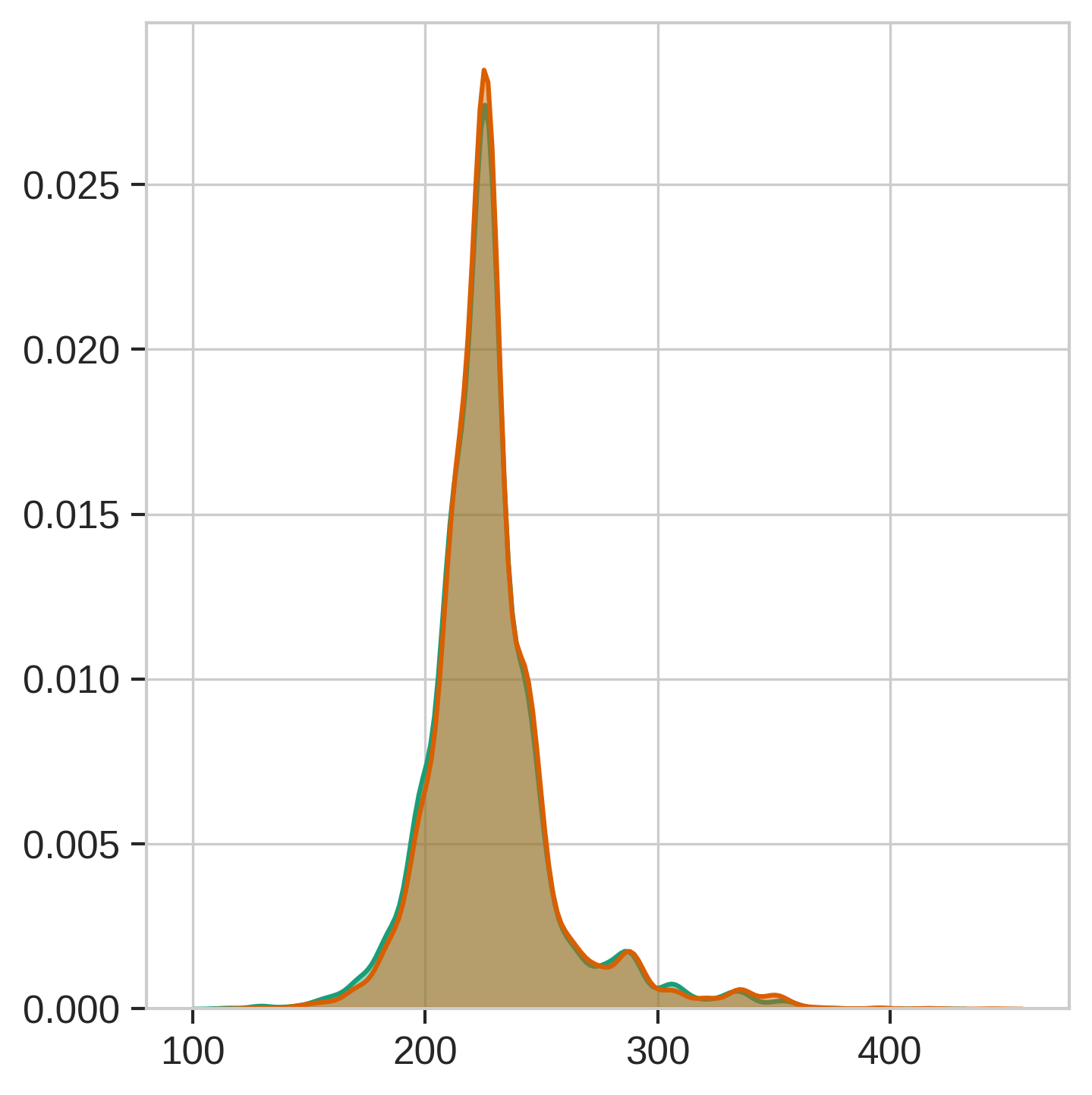} \\
& \small{(D) H-Bond Donors} & \small{(E) Heavy Atom Count} & \small{(F) TPSA} \\
&
\includegraphics[width=0.30\textwidth]{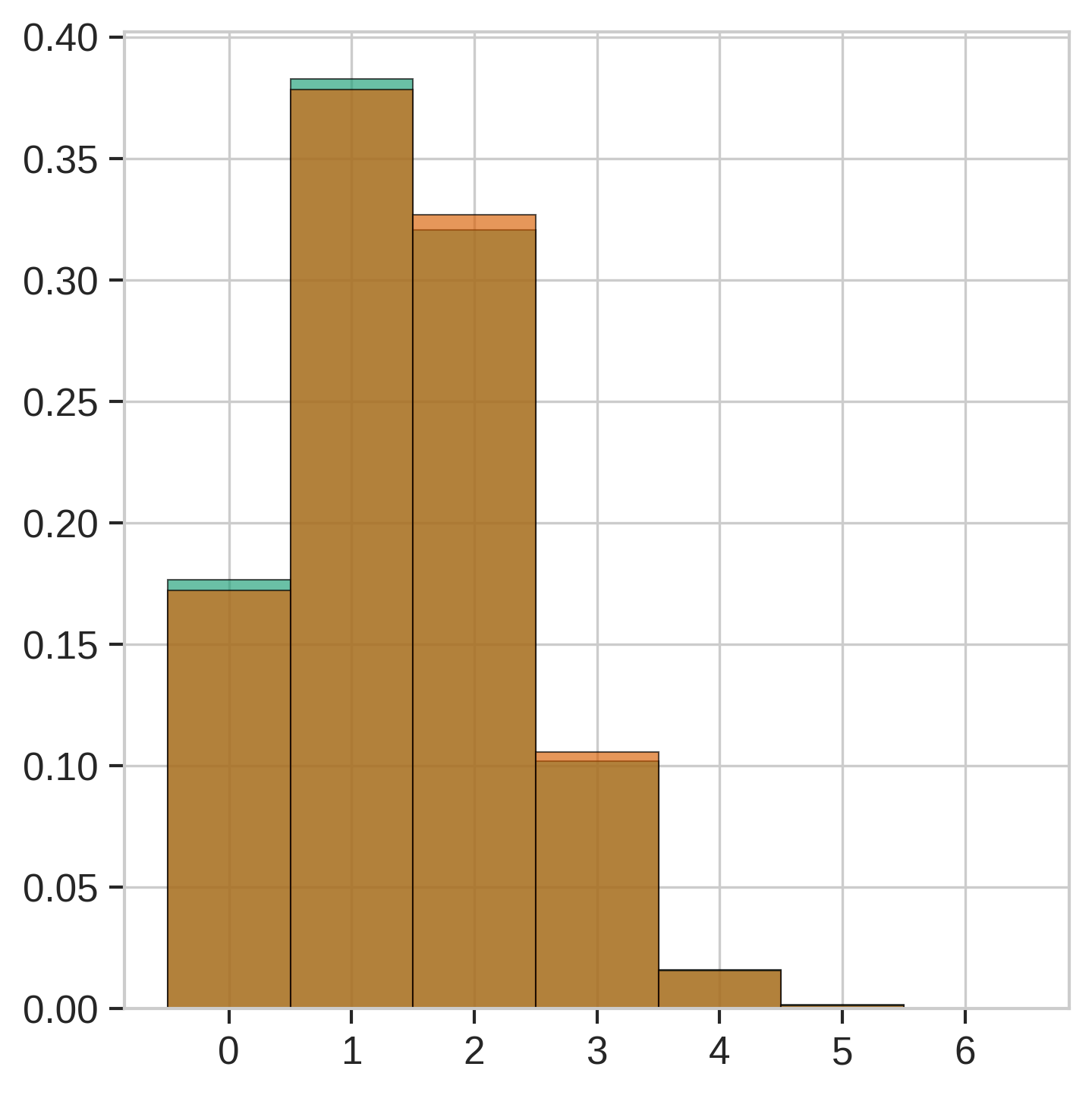} &
\includegraphics[width=0.30\textwidth]{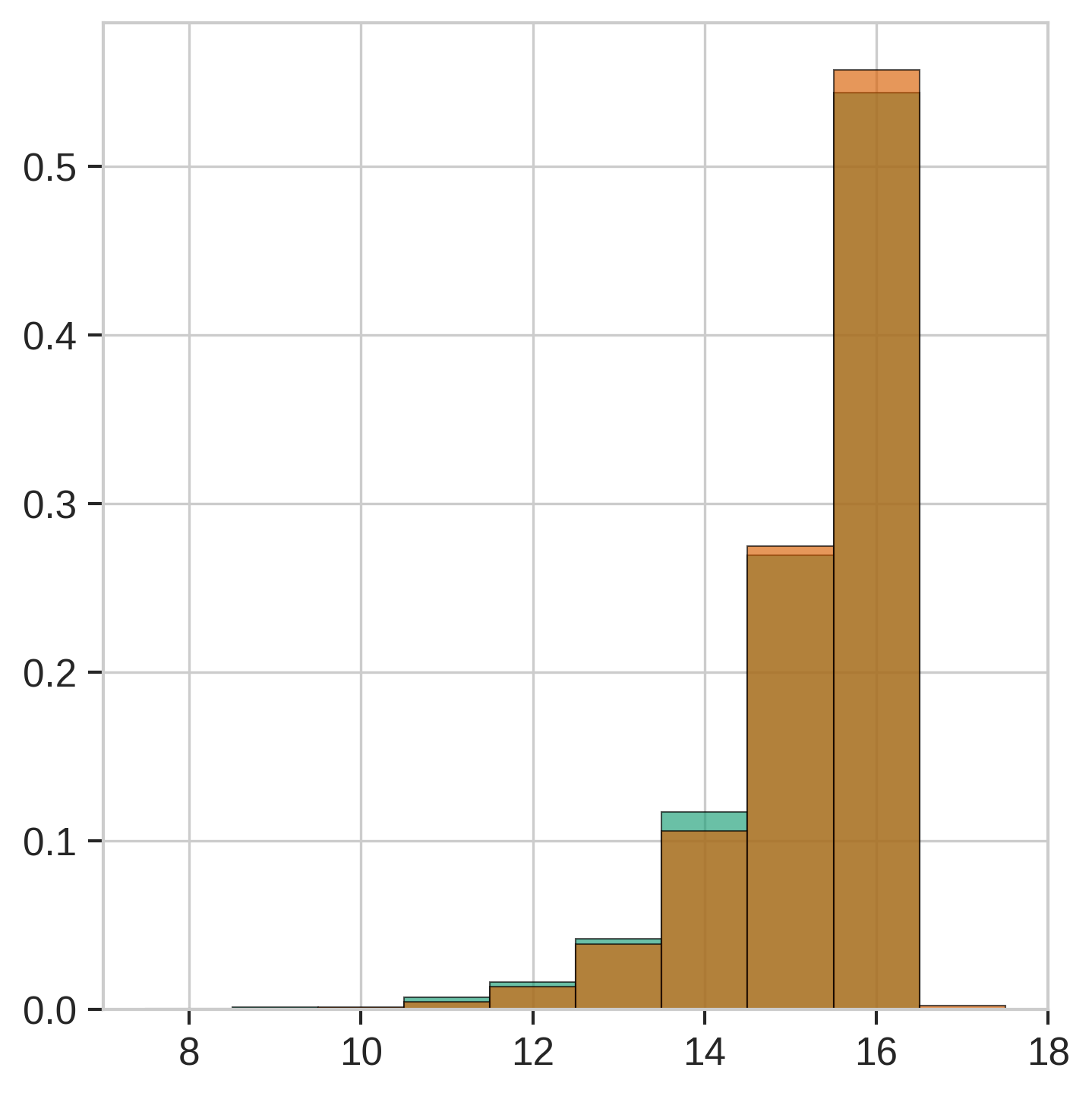} &
\includegraphics[width=0.30\textwidth]{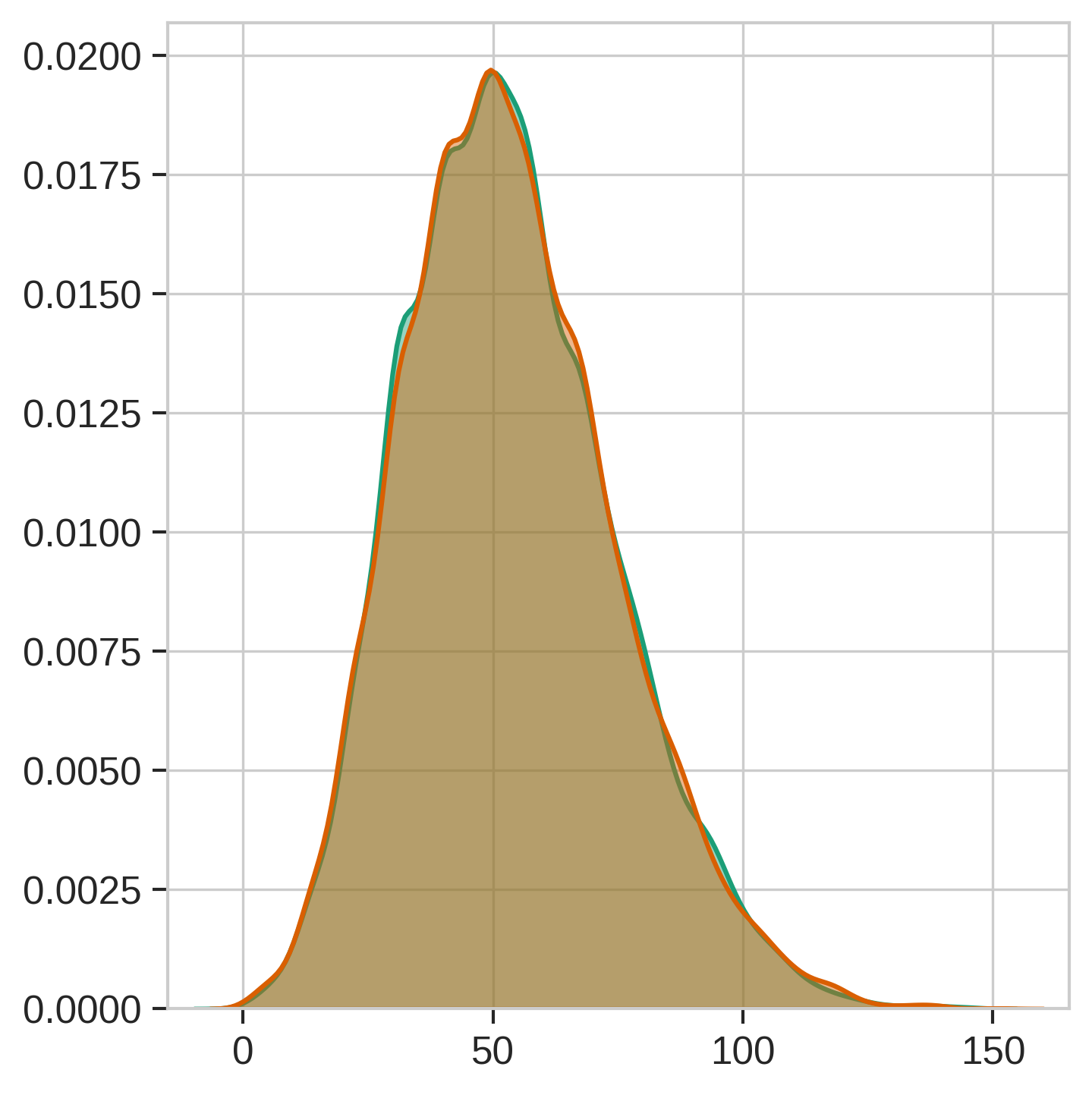}
\end{tabular}
\caption{Distribution plots for six molecular properties comparing generated (blue) and training (orange) molecules: (A) QED, (B) LogP, (C) Molecular weight, (D) H-bond donors, (E) Heavy atom count, (F) TPSA.}
\label{fig:property_distributions}
\end{figure}

\section{Model Architecture and Training}
\begin{table}[b]
\caption{Performance metrics}
\label{tab:validity_diversity}
\centering
\begin{tabular}{lcccccc}
\toprule
\textbf{Metric} & \textbf{Total molecules} & \textbf{Valid molecules} & \textbf{Novel molecules} & \textbf{ZINC coverage} \\
\midrule
Count & 62,015,589 & 61,951,924 & 9,387,465 & 33,210,363 \\
Percentage & 100.00\% & 99.90\% & 22.04\% & 53.55\% \\
\bottomrule
\end{tabular}
\end{table}
\subsection{Dataset}

The model was trained on the complete fragments subset of the ZINC-22 database (April 2025 snapshot), comprising 62,015,589 molecules after preprocessing \cite{tingle2023zinc}. This ZINC-22 2D fragments subset is defined as all 2D molecules with heavy atom count from H08 to H16 (inclusive) and LogP from M500 to P240 (inclusive). In ZINC notation, "H08" through "H16" indicate heavy atom count bins (e.g., H08 = 8 heavy atoms, H16 = 16 heavy atoms), while LogP bins use a signed three-character code where "M" denotes a negative bin and "P" a positive bin (e.g., M500 corresponds to LogP $\leq -5.00$, P240 corresponds to LogP $\geq 2.40$); we use these bin codes when referring to the ZINC fragment subset. Exact string duplicates were removed, but SMILES were not canonicalized during training in order to preserve representational diversity; canonicalization was applied only for evaluation. The release dataset represents roughly 2 billion SMILES tokens with average length 31.54 ± 5.85 characters (range: 8--86), and an interquartile length of 27--35 characters.

\subsection{Architecture}

We used HuggingFace Transformers to develop a GPT-2 based chemical language model tailored for fragment generation. The model uses 6 transformer layers, 12 attention heads, and 768-dimensional embeddings for a total of 42.7M parameters with a 128 token context window, balancing efficiency and capacity for fragment-level SMILES modeling.

\subsection{Training and Implementation}
\label{subsec:implementation}

Training used HuggingFace Transformers on a PyTorch backend with a character-level tokenizer (vocab=42). FragAtlas-62M was trained for 5 epochs on a 99:1 train:validation split with an initial learning rate of 5e-5, a linear schedule with 10\% warmup, and no restarts. Tokenization preserved non-canonical SMILES to leverage augmentation benefits \cite{skinnider2024invalid}; generated SMILES were validated with RDKit sanitization \cite{landrum2006rdkit} and canonicalized for evaluation. The system sustains high-throughput generation (>1,000 molecules/s on a single RTX 4090 GPU) and both training and inference can be performed on consumer-grade GPUs.

\begin{table}[b]
\caption{Molecular property comparison across different analysis sets}
\label{tab:property_comparison}
\centering
\begin{tabular}{lccc}
\toprule
\textbf{Property} & \textbf{Complete} & \textbf{Novel Only} & \textbf{Rediscovered Only} \\
\textbf{} & \textbf{Cohen's d} & \textbf{Cohen's d} & \textbf{Cohen's d} \\
\midrule
QED score & 0.001 & -0.009 & -0.012 \\
LogP & 0.009 & 0.112 & -0.013 \\
H-bond donors & 0.008 & 0.025 & 0.010 \\
TPSA & 0.022 & -0.009 & 0.016 \\
Molecular weight & 0.033 & 0.092 & -0.002 \\
Heavy atom count & 0.053 & 0.099 & 0.014 \\
SA score & 0.042 & 0.121 & -0.001 \\
H-bond acceptors & -0.094 & 0.018 & -0.002 \\
Rotatable bonds & 0.069 & -0.096 & 0.044 \\
Ring count & -0.048 & 0.120 & -0.051 \\
NP score & 0.067 & 0.074 & 0.001 \\
\bottomrule
\end{tabular}
\end{table}

\section{Results and Evaluation}

\subsection{Model Performance}

We generated 62,015,589 SMILES with FragAtlas-62M (primary counts are in Table~\ref{tab:validity_diversity}). Of these, 61,951,924 were chemically valid (99.90\% of all generated SMILES). After deduplication and canonicalization the valid set contains 42,597,827 unique canonical SMILES (42,742,090 unique prior to canonicalization; 19,209,834 exact-string duplicates in the raw outputs; total duplicated entries = 19,354,097). Comparing canonical sets, 33,210,363 molecules are shared with the ground-truth ZINC collection while 9,387,465 generated molecules are novel. The latter equals 46.39\% of all generated SMILES (counting duplicates) or 22.04\% of unique canonical generated SMILES, while coverage of the ZINC fragment set is 53.55\% (ground-truth denominator). Note these percentages use different bases (total generated, unique canonical generated, or the original ZINC fragments set) and are therefore not intended to sum. Across 12 molecular descriptors the effect sizes between generated and training distributions are negligible (all |d| < 0.2; Table~\ref{tab:property_comparison}). 

\begin{figure}[t]
\centering
\begin{tabular}{ccc}
\small{(A) Morgan Fingerprints} & \small{(B) MACCS Keys} & \small{(C) Topological Fingerprints} \\
\includegraphics[width=0.30\textwidth]{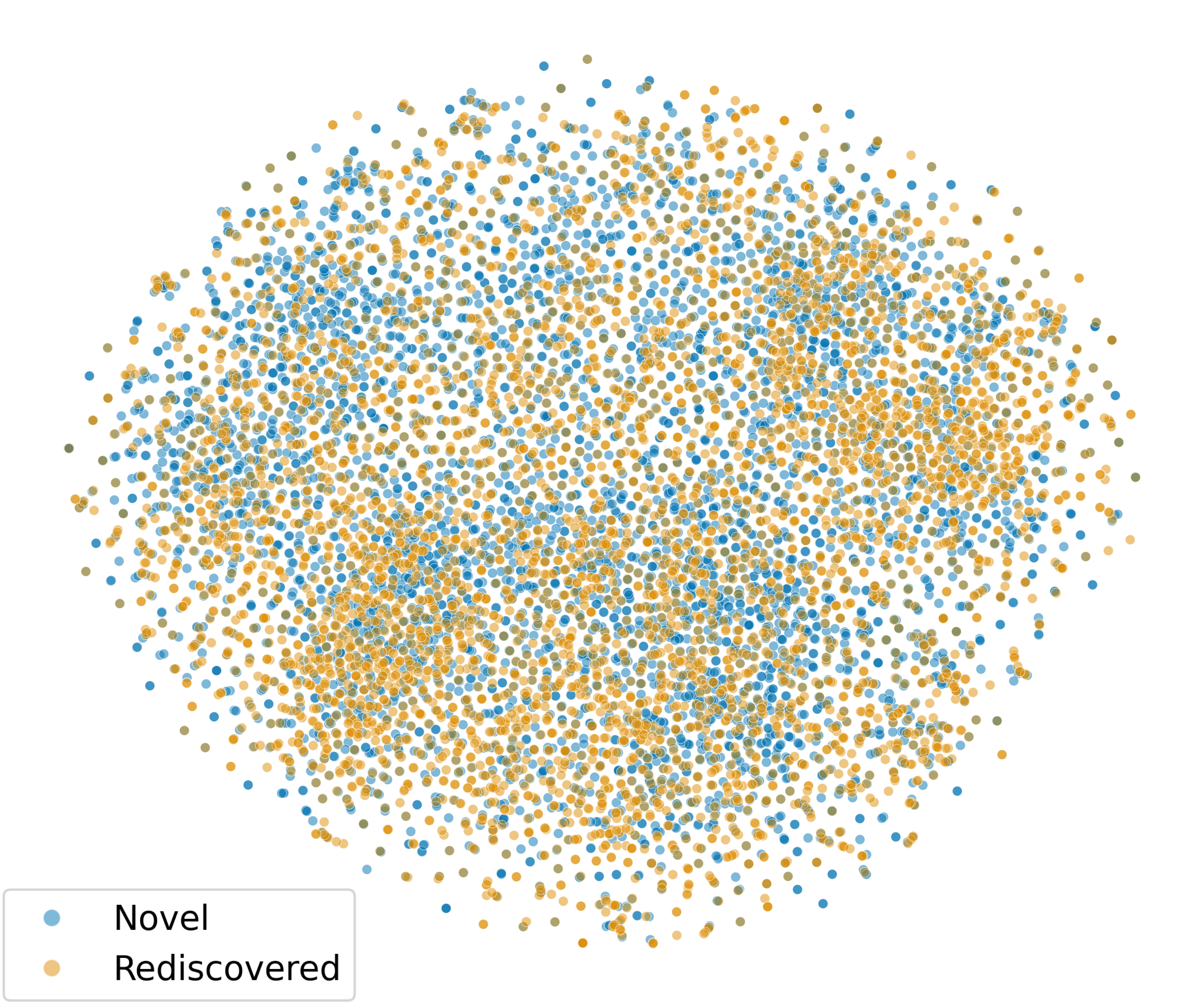} &
\includegraphics[width=0.30\textwidth]{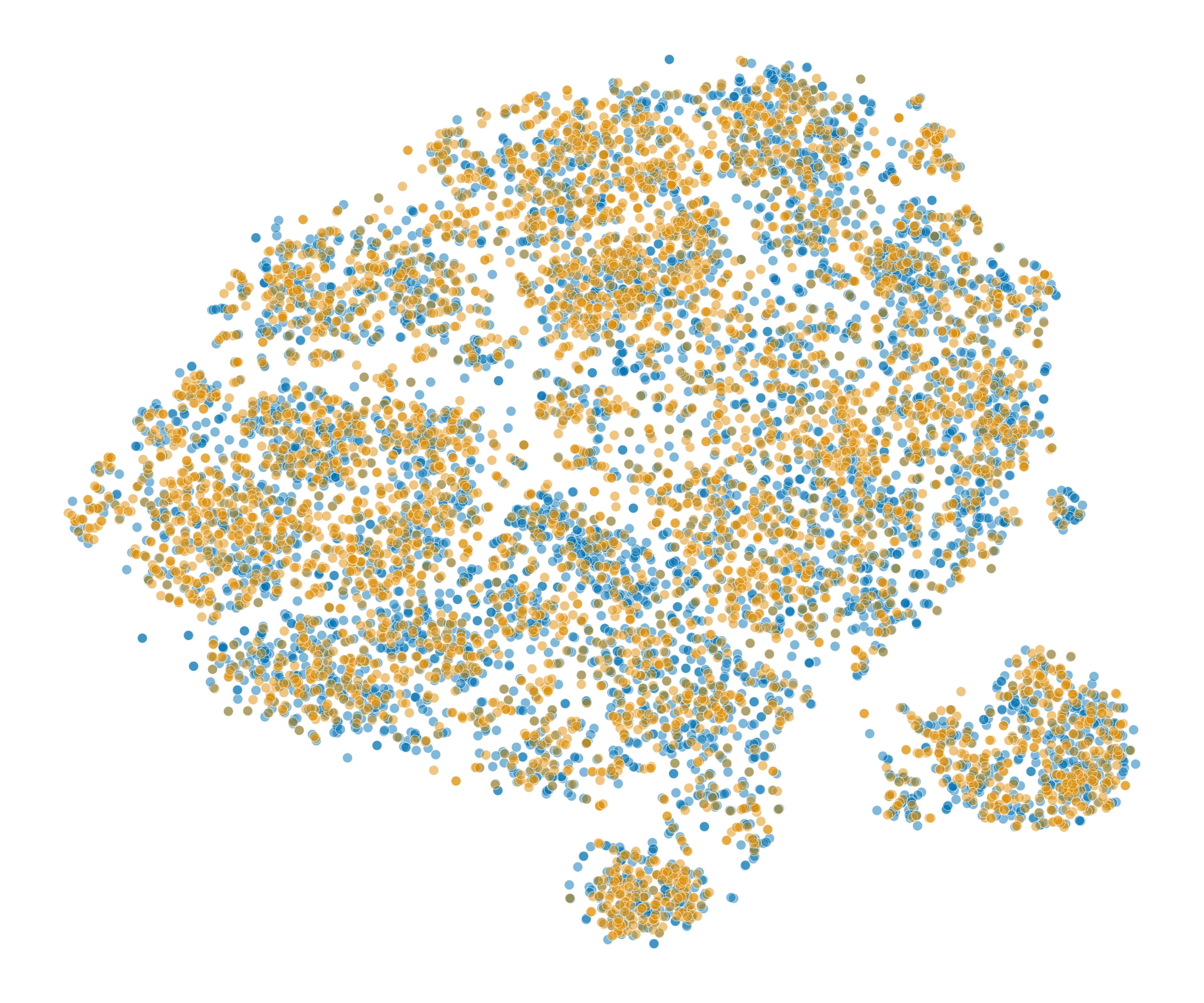} &
\includegraphics[width=0.30\textwidth]{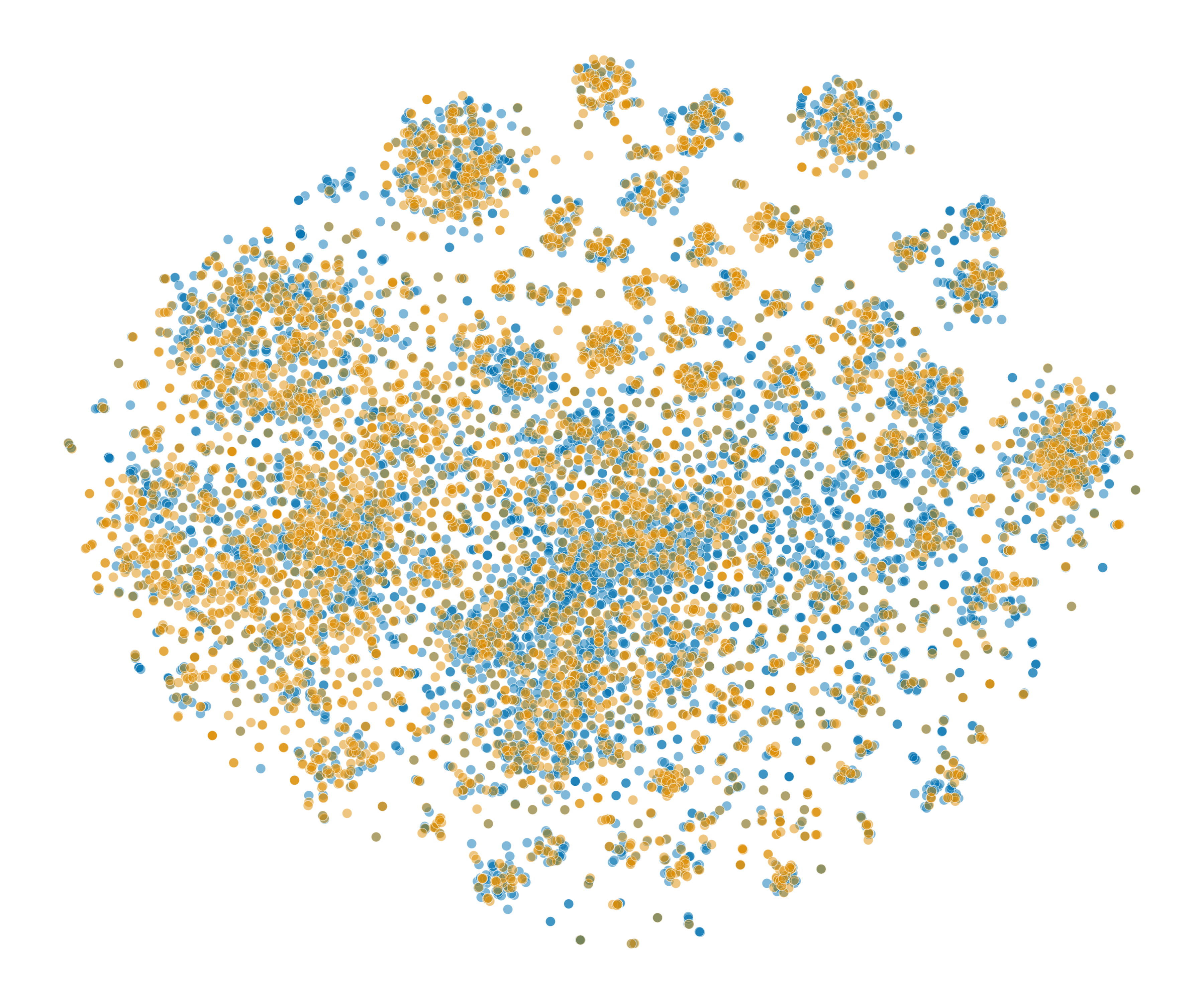}
\end{tabular}
\caption{t-SNE projections of novel (blue) and rediscovered (orange) molecules for three fingerprint types: (A) Morgan (ECFP4), (B) MACCS, (C) Topological.}
\label{fig:fingerprint_tsne}
\end{figure}

\subsection{Chemical Fingerprint Analysis}

To further validate the structural similarity between generated and reference molecules, we performed comprehensive chemical fingerprint analysis. We analyzed novel versus rediscovered molecules using three distinct fingerprint types: Morgan fingerprints (ECFP4, radius=2, 2048 bits), MACCS keys (166 bits), and topological fingerprints (2048 bits). Visualizations of the t-SNE projections for each fingerprint type are shown in Figure~\ref{fig:fingerprint_tsne}.

The distance analysis results, summarized in Table~\ref{tab:fingerprint_analysis}, show distance ratios (NR/NN and NR/RR) consistently near 1.0 across all three fingerprint types. For Morgan fingerprints, the distance ratios were 0.9934 (NR/NN) and 1.0092 (NR/RR). For MACCS keys, the ratios were 0.9934 (NR/NN) and 1.0099 (NR/RR). For Topological fingerprints, the ratios were 0.9780 (NR/NN) and 1.0265 (NR/RR). All effect sizes in group comparisons remained below 0.4, with values ranging from 0.05 to 0.35.

The t-SNE visualizations in Figure~\ref{fig:fingerprint_tsne} show the distributions of novel and rediscovered molecules for each fingerprint type, with substantial overlap observed in all three methods.

\section{Discussion}

FragAtlas-62M demonstrates reliable fragment-level generation that closely follows the training distribution while introducing meaningful novelty. At scale, the model produces chemically valid SMILES (99.90\%), rediscovers more than half of the ZINC fragment set (53.55\%), and generates a substantial fraction of novel fragments (22.04\%). Quantitative comparisons of molecular descriptors (Table~\ref{tab:property_comparison}) and fingerprint analyses (Table~\ref{tab:fingerprint_analysis}) show minimal distributional shifts and substantial overlap in chemotype space, indicating the model generalizes fragment chemistry without systematic bias.

This combination of fidelity and novelty has immediate practical value. FragAtlas-62M generates molecules at high throughput (>1,000 molecules/second on an RTX 4090) and can be fine-tuned or conditioned for target-specific fragment proposals or property-driven design. The provided training code and preprocessed dataset lower the barrier to reuse and extension for groups with modest compute resources.

There are clear limitations to note. FragAtlas-62M does not explicitly model stereochemistry, geometric relationships required for structure-based assembly, or fragment-to-fragment connectivity rules needed for automated lead construction. These gaps limit applications requiring precise 3D control or direct fragment linking for lead optimization \cite{chen2021fragment, irwin2022chemformer}. While the model preserves distributional properties, targeted performance on downstream tasks will require task-specific fine-tuning or integration with structure- or synthesis-aware modules.

Future work should prioritize integrating conditional controls and structural information: adding conditioning inputs, coupling the language model with 3D scoring or docking tools, and developing methods that learn fragment-assembly grammars for automated molecule construction \cite{podda2020deep, lee2024molecule}. Improving stereochemical handling and incorporating synthetic-route awareness would also broaden practical utility.

Overall, FragAtlas-62M provides a practical, extensible foundation for fragment-based discovery: it balances rediscovery and novelty, preserves key properties of the training set, and ships with the data and code needed for rapid follow-up work and experimental validation.

\begin{table}[t]
\caption{Molecular fingerprint distance analysis across different fingerprint types}
\label{tab:fingerprint_analysis}
\centering
\begin{tabular}{lccc}
\toprule
\textbf{Metric} & \textbf{Morgan} & \textbf{MACCS} & \textbf{Topological} \\
\midrule
NN (Intragroup) & 7.4779 ± 0.3995 & 6.3841 ± 0.7692 & 23.7115 ± 3.1989 \\
RR (Intragroup) & 7.3608 ± 0.4250 & 6.2800 ± 0.7821 & 22.5910 ± 3.1784 \\
NR (Intergroup) & 7.4286 ± 0.4111 & 6.3419 ± 0.7735 & 23.1889 ± 3.1927 \\
\midrule
NR/NN Ratio & 0.9934 & 0.9934 & 0.9780 \\
NR/RR Ratio & 1.0092 & 1.0099 & 1.0265 \\
\midrule
\textbf{Effect Sizes (n=1000)} & & & \\
NN vs NR & $d$=0.12 & $d$=0.05 & $d$=0.16 \\
RR vs NR & $d$=0.16 & $d$=0.08 & $d$=0.19 \\
NN vs RR & $d$=0.28 & $d$=0.13 & $d$=0.35 \\
\bottomrule
\multicolumn{4}{l}{\small $d$ = Cohen's d effect size} \\
\multicolumn{4}{l}{\small NN = Novel-Novel, RR = Rediscovered-Rediscovered, NR = Novel-Rediscovered}
\end{tabular}
\end{table}

\bibliographystyle{unsrt}
\bibliography{bib}

\end{document}